\documentclass[hidelinks]{article}
\usepackage{doi}
\usepackage{float}
\usepackage{setspace}
\usepackage[a4paper,margin=1in]{geometry}

\usepackage{amsmath,amssymb,amsfonts}
\usepackage{bm}

\usepackage{graphicx}
\usepackage{caption}
\usepackage{subcaption}
\usepackage{booktabs}
\usepackage{tabularx}
\usepackage{array}
\newcolumntype{L}[1]{>{\raggedright\arraybackslash}p{#1}}
\usepackage{algorithmic}
\usepackage{bbm}

\usepackage{cite}
\usepackage{xcolor}
\usepackage{textcomp}
\usepackage[numbers]{natbib}

\usepackage{authblk}
\usepackage{setspace}
\usepackage{xcolor}
\usepackage{caption}

\definecolor{reviewerblue}{RGB}{0,70,180}

\title{Beyond Accuracy: Uncertainty-Guided Boundary Refinement for Reliable Biomedical Image Segmentation}
\author{Anima Kujur\thanks{Corresponding author}}

\affil{Interdisciplinary Centre for Scientific Computing (IWR),
Heidelberg University, Heidelberg, Germany}

\affil{\texttt{anima.kujur@iwr.uni-heidelberg.de}}
\date{}
\begin{document}
\maketitle

\begin{abstract}
Accurate biomedical image segmentation requires not only high global overlap but also reliable delineation of clinically meaningful boundaries. In blood-smear microscopy, cytoplasm and nucleus contours provide the structural basis for downstream morphology analysis; however, deep segmentation models may remain uncertain or overconfident near ambiguous boundary regions even when achieving strong Dice scores. This work proposes a Reliability-Aware Boundary Refinement Network (RABR-Net), a two-stage framework for reliability-aware biomedical image segmentation. A strong UNet++ EfficientNet-B4 base segmenter first produces initial class probabilities and logits. Predictive entropy, test-time augmentation variance, margin uncertainty, foreground probability gradients, and foreground soft-boundary cues are combined with the multiclass probabilities into a reliability-aware representation. This representation guides a gated residual refiner that selectively corrects uncertain boundary regions while preserving confident regions of the base prediction. The framework is evaluated using overlap accuracy, class-wise Dice, Boundary Dice, HD95/ASSD, expected calibration error, risk–coverage analysis, qualitative correction maps, and paired statistical testing. As a separate diagnostic analysis, we additionally characterize the sensitivity of the selected base segmenter to deterministic microscopy image perturbations.
\end{abstract}

\section{Introduction}
\label{sec:introduction}

White blood cell (WBC) morphology plays an important role in hematological analysis, immune-status assessment, and the screening of abnormal blood-smear patterns~\cite{deshpande2021review}. In routine microscopy, clinicians and laboratory specialists examine WBC appearance to assess cellular characteristics such as cell size, cytoplasmic extent, nuclear shape, nuclear lobulation, and nucleus-to-cytoplasm ratio~\cite{deshpande2021review}. These morphological cues can support downstream tasks such as WBC classification~\cite{sampathila2022customized}, abnormal-cell detection, leukemia-related screening~\cite{matek2019humanlevel}, and treatment monitoring. Accurate segmentation of WBC components is therefore not merely a computer-vision problem; it provides the structural foundation for quantitative morphology analysis in clinically relevant workflows.

A key challenge in WBC segmentation is that biologically meaningful information is concentrated at the boundaries of the cytoplasm and nucleus. Small contour errors may have limited effect on global overlap metrics such as Dice score, but they can substantially affect downstream measurements of nuclear area, cytoplasmic area, nuclear irregularity, and nucleus-to-cytoplasm ratio~\cite{kervadec2019boundary,sudre2017generalized}. This is particularly important for stained blood-smear images, where cytoplasm boundaries may be weak, nuclear contours may be irregular or multilobed, and staining or illumination variations may introduce ambiguity~\cite{kouzehkanan2021segmentation}. Thus, a segmentation model that achieves high global Dice may still be unreliable for morphology-sensitive analysis if it fails at fine boundary regions.

Deep learning-based segmentation methods, particularly U-Net variants and encoder--decoder architectures~\cite{ronneberger2015unet,zhou2018unetpp}, have achieved strong performance in biomedical image segmentation. However, these models are often optimized primarily for global region overlap and may remain overconfident in ambiguous pixels~\cite{guo2017calibration,nguyen2015easilyfooled}. In WBC segmentation, such ambiguity frequently occurs at cytoplasm-background interfaces and nucleus-cytoplasm boundaries. Standard segmentation pipelines usually output a final mask without explicitly indicating where the model is uncertain or whether the predicted boundary is reliable~\cite{gal2016dropout,mehrtash2020confidence}. As a result, boundary errors may be difficult to detect, interpret, or correct.

Most existing WBC segmentation studies emphasize improvements in Dice, IoU, or architectural complexity~\cite{kouzehkanan2021segmentation,kouzehkanan2022raabin}. While these metrics are important, they do not fully capture whether the segmentation is trustworthy. In particular, a segmentation system intended for morphology-sensitive biomedical analysis should answer not only ``what is the predicted mask?'', but also ``where is the model uncertain?''~\cite{gal2016dropout,lakshminarayanan2017ensembles}, ``are boundary errors localized?''~\cite{kervadec2019boundary}, and ``does refinement correct biologically meaningful contour errors without damaging confident regions?'' These questions motivate a reliability-aware view of WBC segmentation, where uncertainty, boundary quality, calibration, and interpretability are considered alongside a separate perturbation-sensitivity analysis of the base segmenter~\cite{mehrtash2020confidence,geifman2017selective}.

To address this problem, we propose a Reliability-Aware Boundary Refinement Network (RABR-Net) for reliability-aware WBC segmentation. The framework is designed as a two-stage segmentation-refinement pipeline. First, a strong UNet++~\cite{zhou2018unetpp} EfficientNet-B4~\cite{tan2019efficientnet} base segmenter produces initial cytoplasm and nucleus predictions. Second, uncertainty and boundary-derived cues~\cite{gal2016dropout,wang2019aleatoric,kervadec2019boundary} are combined into a boundary-aware reliability representation. This representation guides a gated residual refiner~\cite{oktay2018attentionunet} that selectively corrects uncertain boundary pixels while preserving confident regions of the base prediction. Rather than replacing the entire segmentation mask, the refiner learns local residual corrections controlled by a spatial gate, making the refinement process more interpretable and boundary-focused.

The main contributions of this work are as follows:
\begin{itemize}
    \item A WBC-specific reliability-aware segmentation framework that explicitly targets uncertainty and boundary errors in cytoplasm and nucleus segmentation.

    \item A boundary reliability representation combining predictive entropy~\cite{gal2016dropout}, test-time augmentation variance~\cite{wang2019aleatoric}, margin uncertainty~\cite{hendrycks2017baseline}, probability-gradient information, and soft boundary cues~\cite{kervadec2019boundary} to localize ambiguous boundary regions.

    \item A gated residual refinement module~\cite{oktay2018attentionunet} that selectively corrects uncertain boundary pixels instead of replacing the full segmentation, thereby preserving confident regions while improving boundary-sensitive predictions.

    \item A trustworthiness-focused evaluation protocol including global accuracy metrics, class-wise Dice~\cite{milletari2016vnet,sudre2017generalized}, Boundary Dice, HD95/ASSD~\cite{kervadec2019boundary}, calibration metrics~\cite{guo2017calibration,mehrtash2020confidence}, risk--coverage analysis~\cite{geifman2017selective,geifman2019selectivenet}, a separate base-segmenter perturbation-sensitivity analysis under microscopy image perturbations~\cite{hendrycks2017baseline}, qualitative correction maps, and paired statistical testing.
\end{itemize}

Overall, this work shifts the focus from purely maximizing global segmentation overlap toward trustworthy boundary-aware WBC segmentation. The proposed framework is intended to support more reliable morphology-preserving segmentation, where the model's corrections can be spatially inspected through uncertainty maps, refiner gates, and correction maps.

\section{Related Work}
\label{sec:related_work}

\subsection{Deep Learning Architectures for Biomedical Image Segmentation}

Encoder--decoder architectures with skip connections have become the de facto standard for biomedical image segmentation since the introduction of U-Net~\cite{ronneberger2015unet}, which combines a contracting path for context extraction with a symmetric expanding path for precise localization. UNet++~\cite{zhou2018unetpp} extends this design with nested, dense skip pathways that reduce the semantic gap between encoder and decoder feature maps, consistently improving segmentation accuracy across nodule, nuclei, liver, and polyp segmentation tasks. Attention U-Net~\cite{oktay2018attentionunet} introduces attention gates that allow the network to suppress irrelevant regions and focus on target structures without requiring explicit localization modules. V-Net~\cite{milletari2016vnet} was among the first works to combine a U-Net-style architecture with a Dice-based objective specifically designed for volumetric, class-imbalanced segmentation. More recently, transformer-based hybrids such as TransUNet~\cite{chen2021transunet} incorporate global self-attention into the encoder to capture long-range dependencies while retaining CNN-based decoders for fine-grained localization. Beyond individual architectures, nnU-Net~\cite{isensee2021nnunet} demonstrated that a self-configuring U-Net pipeline, without manual architectural search, can outperform specialized solutions across a wide range of biomedical segmentation benchmarks, underscoring that architecture choice alone is often less critical than proper configuration and training strategy. Litjens et al.~\cite{litjens2017survey} provide a comprehensive survey of these developments, tracing the field's shift from handcrafted features to fully learned representations. Our base segmenter builds on this lineage, adopting a UNet++ decoder with an EfficientNet-B4~\cite{tan2019efficientnet} encoder pretrained via standard backbone training and optimized with AdamW~\cite{loshchilov2019decoupledweightdecayregularization}; unlike prior work, however, our focus is not on proposing a new backbone but on characterizing and correcting the reliability of its boundary predictions.

\subsection{Loss Functions for Class-Imbalanced and Boundary-Sensitive Segmentation}

Standard region-based losses such as cross-entropy and Dice loss are known to be dominated by large, easy regions, leading to degraded performance on small or thin structures such as nuclear and cytoplasmic boundaries. The Generalized Dice loss~\cite{sudre2017generalized} addresses class imbalance by re-weighting the contribution of each class according to its inverse volume, improving stability when foreground structures occupy a small fraction of the image. Kervadec et al.~\cite{kervadec2019boundary} instead propose a boundary loss that operates directly on the space of contours rather than regions, complementing regional losses and yielding measurable improvements in boundary-sensitive metrics such as the Hausdorff distance. Focal loss~\cite{lin2017focal}, originally proposed for dense object detection, similarly down-weights well-classified examples to concentrate learning capacity on harder, more ambiguous pixels -- a property directly relevant to cytoplasm-background and nucleus-cytoplasm interfaces. These boundary- and imbalance-aware objectives motivate our use of boundary-derived cues (probability gradients, soft boundary maps) as explicit inputs to the reliability representation that guides our refinement stage, rather than relying solely on the base segmenter's regional loss to resolve boundary ambiguity.

\subsection{Uncertainty Quantification in Deep Neural Networks}

Estimating predictive uncertainty is central to building trustworthy segmentation systems. Gal and Ghahramani~\cite{gal2016dropout} showed that dropout applied at test time can be interpreted as an approximation to Bayesian inference, enabling cheap epistemic uncertainty estimates from an otherwise deterministic network. Lakshminarayanan et al.~\cite{lakshminarayanan2017ensembles} proposed deep ensembles as a simple, well-calibrated alternative that captures model uncertainty through the disagreement of independently trained networks, and further demonstrated that such uncertainty increases meaningfully on out-of-distribution inputs. Complementary to model (epistemic) uncertainty, Wang et al.~\cite{wang2019aleatoric} introduced a test-time augmentation framework for estimating aleatoric, image-dependent uncertainty in medical image segmentation, formalizing it as Monte Carlo simulation over an image acquisition model with stochastic transformations. For detecting unreliable or out-of-distribution predictions, Hendrycks and Gimpel~\cite{hendrycks2017baseline} established the widely used softmax-confidence baseline, showing that correctly classified and in-distribution examples tend to receive higher maximum softmax probabilities than misclassified or anomalous ones. Our framework draws on this body of work by combining predictive entropy, test-time augmentation variance, and margin-based uncertainty into a unified boundary-aware reliability signal, rather than treating any single uncertainty estimator in isolation.

\subsection{Calibration of Segmentation Models}

A well-calibrated model is one whose predicted confidence reflects the true likelihood of correctness. Guo et al.~\cite{guo2017calibration} demonstrated that modern deep networks are systematically overconfident and showed that simple post-hoc methods such as temperature scaling can substantially reduce calibration error on classification tasks. Mehrtash et al.~\cite{mehrtash2020confidence} extended this analysis specifically to fully convolutional segmentation networks, showing that U-Net-style architectures trained with Dice loss are similarly miscalibrated, and proposing model ensembling and evaluation protocols tailored to pixel-wise predictive uncertainty and out-of-distribution detection in medical imaging. These findings motivate our explicit evaluation of expected calibration error (ECE) alongside overlap and boundary accuracy, and our examination of whether improved segmentation geometry is accompanied by improved probability calibration.

\subsection{Selective Prediction and Risk--Coverage Analysis}

Selective prediction, or the reject option, allows a model to abstain on inputs where its confidence is low, trading coverage for reliability. Geifman and El-Yaniv~\cite{geifman2017selective} formalized this trade-off for deep neural networks, proposing a method to construct a selective classifier that guarantees a target risk level with high probability. Their follow-up work, SelectiveNet~\cite{geifman2019selectivenet}, integrates the reject option directly into network training via an auxiliary selection head, jointly optimizing prediction and abstention. We adapt the risk–coverage framework to pixel-wise semantic segmentation, where pixels are ranked by uncertainty and the classification error among the retained low-uncertainty pixels is evaluated as coverage varies.

\subsection{White Blood Cell Segmentation Datasets and Methods}

Progress in automated leukocyte analysis has been constrained by the scarcity of large, expert-annotated datasets with pixel-level nucleus and cytoplasm labels. The Raabin-WBC dataset~\cite{kouzehkanan2022raabin} addresses this gap by providing approximately 40{,}000 white blood cell images collected across multiple cameras and microscopes, with segmented nuclei and cytoplasm ground truth available for a curated subset, explicitly designed to support classification, detection, and segmentation research. Kouzehkanan et al.~\cite{kouzehkanan2021segmentation} further proposed a dedicated nucleus segmentation and cytoplasm extraction pipeline evaluated on Raabin-WBC alongside the LISC and BCCD datasets, reporting strong Dice similarity for nucleus segmentation and highlighting the generalization challenges posed by staining and imaging variability across sources. These works establish both the data foundation and the domain-specific difficulty -- ambiguous cytoplasm boundaries, irregular or multilobed nuclear contours, and cross-device variability -- that motivate a reliability-aware, rather than purely accuracy-driven, evaluation of WBC segmentation models.

\section{Data}
\label{sec:data}

We evaluated the proposed framework on the WbcMSBench dataset from the MedSegBench benchmark suite \cite{article}, which provides annotated white blood cell (WBC) microscopy images for semantic segmentation. Each sample consists of an RGB blood-smear image and a corresponding pixel-wise segmentation mask. The task is formulated as a three-class segmentation problem,
\[
\mathcal{C}=\{0,1,2\},
\]
where class $0$ denotes background, class $1$ denotes cytoplasm, and class $2$ denotes nucleus. Thus, for each image
\[
x_i \in [0,1]^{3\times H \times W},
\]
the corresponding segmentation mask is
\[
y_i \in \{0,1,2\}^{H\times W}.
\]
In this study, all images were processed at spatial resolution $H=W=256$.

\subsection{Dataset Splits}
\label{subsec:data_splits}

We used the official train, validation, and test splits provided by WbcMSBench. The training set was used for model optimization, the validation set was used for model selection and hyperparameter decisions, and the held-out test set was used only for final reporting. This separation was maintained for both the base segmentation model and the proposed reliability-aware refinement module. The resulting dataset splits and their roles in model development and evaluation are summarized in Table \ref{tab:data_splits}.

\begin{table}[t]
\centering
\caption{
Dataset splits used in the experiments. The validation set was used for model selection, while the test set was held out for final evaluation.
}
\label{tab:data_splits}
\begin{tabular}{lcc}
\toprule
\textbf{Split} & \textbf{Number of images} & \textbf{Usage} \\
\midrule
Training & 280 & Model optimization \\
Validation & 40 & Model selection and hyperparameter tuning \\
Test & 80 & Final held-out evaluation \\
\bottomrule
\end{tabular}
\end{table}

\subsection{Segmentation Labels}
\label{subsec:segmentation_labels}

The segmentation masks contain three semantic labels. The background class represents non-WBC regions, including surrounding smear content. The cytoplasm class represents the visible cell body, while the nucleus class represents the nuclear region inside the WBC. For a pixel location $u\in\Omega$, the ground-truth label is denoted by $y(u)$. Class-wise binary masks are defined as
\[
Y_c(u)=\mathbbm{1}[y(u)=c],
\qquad c\in\{0,1,2\}.
\]
The foreground WBC region is defined as the union of cytoplasm and nucleus:
\[
Y_{\mathrm{fg}}(u)=\mathbbm{1}[y(u)>0].
\]
This distinction is important because global segmentation accuracy alone may hide clinically relevant boundary errors. Therefore, in addition to global overlap metrics, we evaluate cytoplasm Dice, nucleus Dice, boundary Dice, HD95, and ASSD.

\subsection{Preprocessing}
\label{subsec:data_preprocessing}

Each image was converted to a three-channel RGB tensor and normalized to the range $[0,1]$:
\[
x_i \leftarrow \frac{x_i}{255}.
\]
All images were resized to $256\times256$ pixels using image interpolation, while masks were resized using nearest-neighbor interpolation to preserve integer class labels:
\[
x_i \in [0,1]^{3\times256\times256},
\qquad
y_i \in \{0,1,2\}^{256\times256}.
\]
No intensity normalization was computed from the test set. The same deterministic resizing and label conversion were applied to the training, validation, and test sets.

\subsection{Training-Time Data Augmentation}
\label{subsec:data_augmentation}

Data augmentation was applied only to the training split. The validation and test sets were never augmented. The augmentation strategy was designed to simulate common variability in blood-smear microscopy, including differences in cell orientation, illumination, staining, and local shape deformation. Specifically, the training pipeline included random horizontal and vertical flips, geometric transformations, brightness and contrast perturbations, hue-saturation-value changes, local contrast enhancement, Gaussian blur, Gaussian noise, and elastic deformation.

Let $T$ denote a randomly sampled label-preserving transformation. During training, the model receives
\[
(x_i',y_i') = T(x_i,y_i),
\]
where the same geometric transformation is applied to both image and mask. Photometric transformations are applied only to the image:
\[
x_i' = T_{\mathrm{photo}}(x_i),
\qquad
y_i' = y_i.
\]
This ensures that the segmentation label remains spatially consistent with the transformed image. For validation and testing,
\[
(x_i',y_i')=(x_i,y_i),
\]
except for deterministic resizing.

\subsection{Cached Prediction Data for Refinement}
\label{subsec:cached_prediction_data}

The refinement stage does not retrain the base segmenter. Instead, the base-model outputs and TTA-derived quantities required by the second stage were precomputed and reused during refiner training and evaluation. This design keeps the two stages modular and avoids repeatedly executing the full base segmentation model during refinement experiments.

In the notation introduced formally in Section 4.4, \(\bar{p}\) denotes the TTA-averaged class-probability map and \(\bar{z}_{b}(c,u)=\log(\bar{p}(c,u)+\epsilon)\) denotes the corresponding log representation used by the residual refiner. These quantities, together with the image, target mask, and TTA-derived uncertainty information, form the cached inputs from which the reliability representation is constructed. The complete reliability tensor used by the refiner is defined in Section 4.5. Throughout the remainder of the paper, we refer to the segmentation obtained from the TTA-averaged probability map \(\bar p\) as the cached TTA-mean base prediction.

\subsection{Held-Out Test Protocol}
\label{subsec:heldout_test_protocol}

All model choices were made using the validation set. The held-out test set was used only after the base model, refiner, post-processing settings, and evaluation protocol had been fixed. The final test-set evaluation reports segmentation accuracy, boundary quality, probability calibration, risk–coverage behavior, and paired statistical reliability. Separately, we characterize the sensitivity of the selected base segmenter to deterministic microscopy image perturbations.

\section{Methods}
\label{sec:methods}

\begin{figure}[t]
    \centering
    \includegraphics[width=\linewidth]{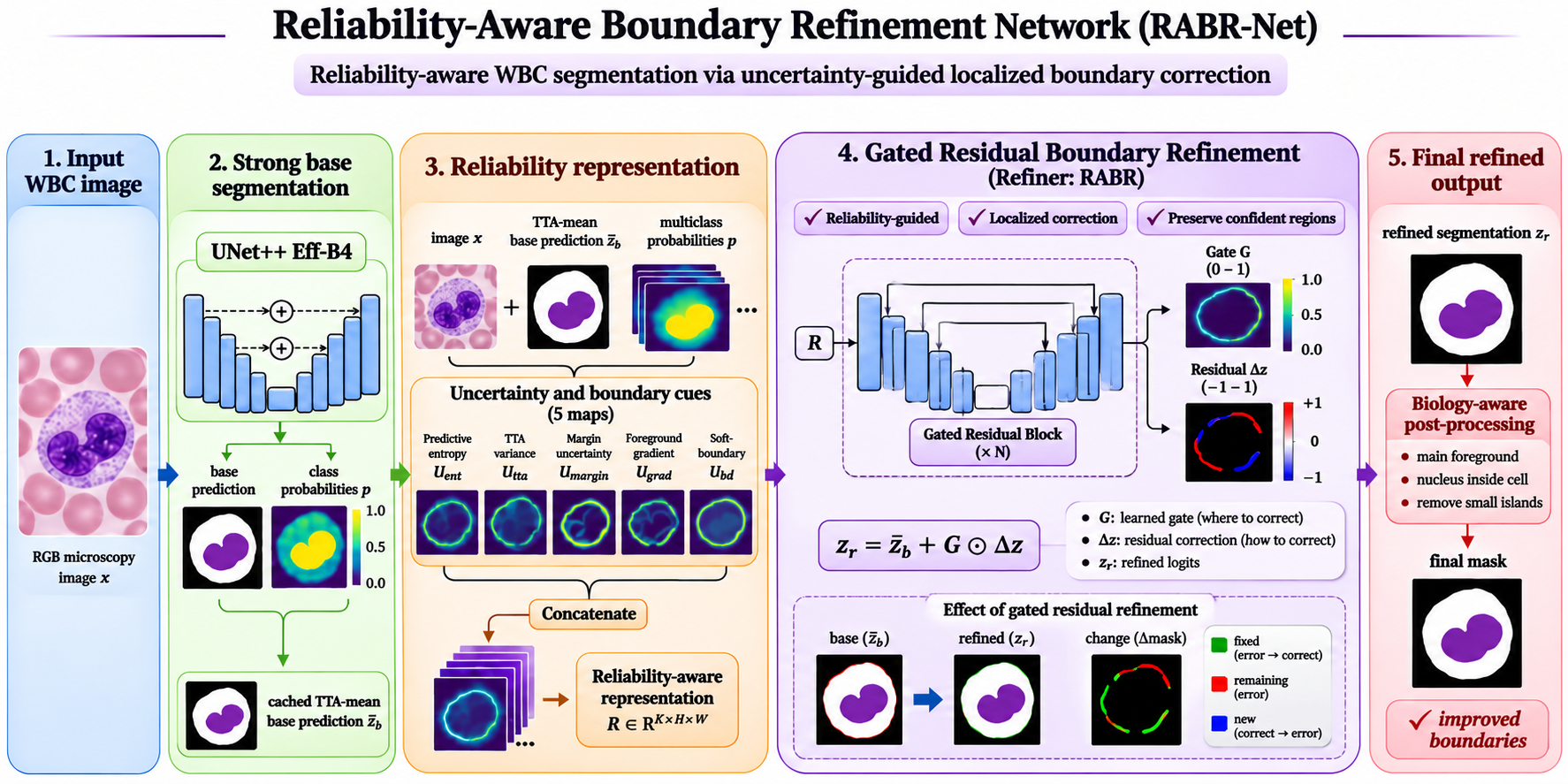}
    \caption{Overview of RABR-Net. A UNet++ EfficientNet-B4 base segmenter provides TTA-averaged predictions and uncertainty/boundary cues, which form a reliability-aware representation. A gated residual refiner selectively corrects ambiguous boundary regions, followed by biology-aware post-processing to produce the final WBC segmentation.
    }
    \label{fig:workflow}
\end{figure}

We propose a Reliability-Aware Boundary Refinement Network (RABR-Net) for reliability-aware white blood cell (WBC) segmentation. The proposed framework is designed as a two-stage segmentation-refinement pipeline. First, a strong base segmenter produces an initial three-class segmentation together with pixel-wise probabilistic outputs. Second, uncertainty and boundary-derived information are combined into a reliability-aware representation that guides a gated residual refinement network. Instead of replacing the entire segmentation mask, the refiner predicts local residual corrections and applies them through a learned spatial gate, allowing the model to selectively correct uncertain cytoplasm and nucleus boundary pixels while preserving confident regions of the base prediction. The full pipeline is illustrated in Fig.~\ref{fig:workflow}.

\subsection{Problem Formulation and Notation}
\label{subsec:problem_formulation}

As described in Section~\ref{sec:data}, each input image is represented as
\[
x_i \in [0,1]^{3 \times H \times W},
\]
with corresponding segmentation mask
\[
y_i \in \{0,1,2\}^{H \times W},
\]
where the three semantic labels correspond to background, cytoplasm, and nucleus. In all experiments, $H=W=256$. The goal is to learn a segmentation function that predicts a dense label map
\[
\hat{y}_i \in \{0,1,2\}^{H \times W}.
\]

The base segmenter produces logits
\[
z_b = f_{\theta}(x_i) \in \mathbb{R}^{C \times H \times W},
\]
where $C=3$ is the number of classes. The corresponding softmax probabilities are
\[
p_b(c,u)
=
\frac{\exp(z_b(c,u))}
{\sum_{k=0}^{C-1}\exp(z_b(k,u))},
\qquad c\in\{0,1,2\},
\]
where $u\in\Omega$ indexes a spatial pixel location. The initial base prediction is
\[
\hat{y}_b(u)
=
\arg\max_{c\in\{0,1,2\}} p_b(c,u).
\]

The proposed refinement stage takes the image, base probabilities, uncertainty maps, and boundary-derived maps as input and predicts a refined segmentation
\[
\hat{y}_r \in \{0,1,2\}^{H \times W}.
\]
The central objective is not only to improve global overlap metrics, but to improve boundary-sensitive reliability by correcting uncertain cytoplasm and nucleus contour regions.

\subsection{Base Segmentation Model}
\label{subsec:base_segmenter}

The primary base segmenter was a UNet++ architecture with an EfficientNet-B4 encoder initialized with ImageNet weights. The network maps an input image $x$ to class logits
\[
z_b = f_{\theta}(x) \in \mathbb{R}^{C \times H \times W},
\]
where $C=3$ is the number of classes. Pixel-wise class probabilities are obtained using the softmax function:
\[
p_b(c,u) = 
\frac{\exp(z_b(c,u))}
{\sum_{k=0}^{C-1}\exp(z_b(k,u))},
\qquad c \in \mathcal{C},
\]
where $u$ indexes a spatial pixel location. The base prediction is then given by
\[
\hat{y}_b(u) = \arg\max_{c \in \mathcal{C}} p_b(c,u).
\]

The base segmenter was trained using AdamW with learning rate $3\times 10^{-4}$ and weight decay $10^{-4}$. A cosine annealing learning-rate schedule was used, with a maximum of 150 epochs and early stopping patience of 25 epochs. The batch size was set to 8. Mixed precision training was enabled when CUDA was available. Gradients were clipped with maximum norm 1.0. The best base model was selected using the validation score
\[
S_{\mathrm{base}} =
\mathrm{Dice}_{\mathrm{val}}
+
0.25\,\mathrm{BoundaryDice}_{\mathrm{val}},
\]
which encourages the selected model to balance global overlap and boundary quality.

\subsection{Compound Segmentation Loss}
\label{subsec:loss}

The base segmenter and the refinement network were trained using a compound segmentation loss consisting of cross-entropy loss, multiclass Dice loss, focal loss, and a boundary Dice loss. The total loss is
\[
\mathcal{L}_{\mathrm{seg}}
=
\lambda_{\mathrm{ce}}\mathcal{L}_{\mathrm{ce}}
+
\lambda_{\mathrm{dice}}\mathcal{L}_{\mathrm{dice}}
+
\lambda_{\mathrm{focal}}\mathcal{L}_{\mathrm{focal}}
+
\lambda_{\mathrm{bd}}\mathcal{L}_{\mathrm{bd}},
\]
with weights
\[
\lambda_{\mathrm{ce}}=0.35,\qquad
\lambda_{\mathrm{dice}}=0.35,\qquad
\lambda_{\mathrm{focal}}=0.10,\qquad
\lambda_{\mathrm{bd}}=0.20.
\]

For a pixel $u$, the cross-entropy loss is
\[
\mathcal{L}_{\mathrm{ce}}
=
-\frac{1}{|\Omega|}
\sum_{u\in\Omega}
\log p(y(u),u),
\]
where $\Omega$ denotes the set of image pixels and $p(y(u),u)$ is the predicted probability of the ground-truth class at pixel $u$.

The multiclass Dice loss is computed over the foreground classes and penalizes mismatch between the predicted soft probabilities and the one-hot encoded ground truth. For class $c$, the soft Dice coefficient is
\[
\mathrm{Dice}_c
=
\frac{
2\sum_{u\in\Omega} p(c,u) \, \mathbbm{1}[y(u)=c] + \epsilon
}{
\sum_{u\in\Omega} p(c,u)
+
\sum_{u\in\Omega} \mathbbm{1}[y(u)=c]
+
\epsilon
},
\]
and the Dice loss is
\[
\mathcal{L}_{\mathrm{dice}}
=
1-\frac{1}{C-1}\sum_{c=1}^{C-1}\mathrm{Dice}_c.
\]

To emphasize difficult pixels, focal loss was also included:
\[
\mathcal{L}_{\mathrm{focal}}
=
-\frac{1}{|\Omega|}
\sum_{u\in\Omega}
\left(1-p(y(u),u)\right)^{\gamma}
\log p(y(u),u),
\]
where $\gamma$ is the focusing parameter used by the segmentation loss implementation.

Because the main weakness of high-performing WBC segmentation models is often boundary accuracy, we included a differentiable boundary Dice loss. For a soft class map $q$, a soft boundary operator is approximated by local max-min pooling:
\[
\mathcal{B}(q)
=
\mathrm{MaxPool}_{3\times 3}(q)
-
\mathrm{MinPool}_{3\times 3}(q).
\]
In implementation, the min-pooling operation is computed as
\[
\mathrm{MinPool}(q)
=
-\mathrm{MaxPool}(-q).
\]
Let $p_{1:C-1}$ denote the foreground probability maps and $Y_{1:C-1}$ the one-hot encoded foreground masks. The predicted and target boundary maps are
\[
B_p = \mathcal{B}(p_{1:C-1}),
\qquad
B_y = \mathcal{B}(Y_{1:C-1}).
\]
The boundary Dice loss is then
\[
\mathcal{L}_{\mathrm{bd}}
=
1-
\frac{1}{C-1}
\sum_{c=1}^{C-1}
\frac{
2\sum_{u\in\Omega} B_p(c,u)B_y(c,u) + \epsilon
}{
\sum_{u\in\Omega}B_p(c,u)
+
\sum_{u\in\Omega}B_y(c,u)
+
\epsilon
}.
\]

\subsection{Uncertainty Estimation}
\label{subsec:uncertainty}

After training the base segmenter, we used its probabilistic output to estimate pixel-wise uncertainty. The uncertainty maps are later used to construct the reliability-aware input to the refinement network.

First, normalized predictive entropy was computed as
\[
U_{\mathrm{ent}}(u)
=
-\frac{1}{\log C}
\sum_{c=0}^{C-1}
p_b(c,u)\log\left(p_b(c,u)+\epsilon\right).
\]
This quantity lies in $[0,1]$, with higher values corresponding to higher uncertainty.

Second, margin uncertainty was computed from the difference between the two largest class probabilities. Let $p_{(1)}(u)$ and $p_{(2)}(u)$ denote the largest and second-largest predicted probabilities at pixel $u$. The margin uncertainty is
\[
U_{\mathrm{margin}}(u)
=
1-\left(p_{(1)}(u)-p_{(2)}(u)\right).
\]
Pixels with similar top-two class probabilities therefore receive higher uncertainty values.

Third, test-time augmentation (TTA) uncertainty was estimated using horizontal and vertical flips. Specifically, we evaluated the base model under four transformations:
\[
\mathcal{T}=\{\mathrm{id}, \mathrm{hflip}, \mathrm{vflip}, \mathrm{hflip}\circ\mathrm{vflip}\}.
\]
For each transformation $t\in\mathcal{T}$, the image was transformed, passed through the base model, and the predicted probability map was transformed back to the original orientation:
\[
p_t = t^{-1}\left(\mathrm{softmax}(f_{\theta}(t(x)))\right).
\]
The mean TTA probability was then computed as
\[
\bar{p}(c,u)
=
\frac{1}{|\mathcal{T}|}
\sum_{t\in\mathcal{T}} p_t(c,u).
\]
The TTA variance map was computed as the mean class-wise variance:
\[
U_{\mathrm{tta}}(u)
=
\frac{1}{C}
\sum_{c=0}^{C-1}
\mathrm{Var}_{t\in\mathcal{T}}\left[p_t(c,u)\right].
\]
This map was normalized image-wise by its maximum value:
\[
U_{\mathrm{tta}}(u)
\leftarrow
\frac{U_{\mathrm{tta}}(u)}
{\max_{v\in\Omega} U_{\mathrm{tta}}(v)+\epsilon}.
\]
For downstream refinement, the base logits were reconstructed from the averaged probabilities as
\[
\bar{z}_b(c,u)=\log\left(\bar{p}(c,u)+\epsilon\right).
\]

\subsection{Boundary-Aware Reliability Map}
\label{subsec:boundary_reliability}

In addition to uncertainty, the method uses boundary evidence to identify spatial regions where refinement is most likely to be useful. The foreground probability is defined as
\[
p_{\mathrm{fg}}(u)
=
\sum_{c=1}^{C-1}\bar{p}(c,u).
\]
A Sobel-like probability-gradient map is computed from the foreground probability:
\[
G_x = K_x * p_{\mathrm{fg}},
\qquad
G_y = K_y * p_{\mathrm{fg}},
\]
where $*$ denotes convolution and
\[
K_x =
\frac{1}{8}
\begin{bmatrix}
-1 & 0 & 1\\
-2 & 0 & 2\\
-1 & 0 & 1
\end{bmatrix},
\qquad
K_y = K_x^{\top}.
\]
The normalized probability-gradient magnitude is
\[
U_{\mathrm{grad}}(u)
=
\frac{
\sqrt{G_x(u)^2+G_y(u)^2+\epsilon}
}{
\max_{v\in\Omega}\sqrt{G_x(v)^2+G_y(v)^2+\epsilon}
+
\epsilon
}.
\]

A soft boundary hint is also computed from the foreground probability using local max-min pooling:
\[
U_{\mathrm{bd}}(u)
=
\mathrm{MaxPool}_{3\times3}\left(p_{\mathrm{fg}}\right)(u)
-
\mathrm{MinPool}_{3\times3}\left(p_{\mathrm{fg}}\right)(u).
\]
Because \(p_{\mathrm{fg}}\) aggregates the cytoplasm and nucleus probabilities, \(U_{\mathrm{grad}}\) and \(U_{\mathrm{bd}}\) primarily encode changes at the outer WBC-to-background contour. Ambiguity at the internal nucleus–cytoplasm interface is represented by the multiclass probability channels together with entropy, margin uncertainty, and TTA-based uncertainty. Thus, the refiner receives complementary information about both outer foreground boundaries and multiclass ambiguity within the cell.

The final reliability tensor used as input to the refiner concatenates the original image, base probabilities, and uncertainty/boundary maps:
\[
R(x)
=
\left[
x,\,
\bar{p},\,
U_{\mathrm{ent}},\,
U_{\mathrm{margin}},\,
U_{\mathrm{tta}},\,
U_{\mathrm{grad}},\,
U_{\mathrm{bd}}
\right].
\]
The individual channels in \(R(x)\), rather than a single collapsed reliability score, constitute the actual input to the residual refiner. Scalar reliability maps introduced below are used only for visualization and qualitative interpretation.

Since $x$ has 3 channels, $\bar{p}$ has 3 channels, and the five reliability maps each have one channel, the resulting refiner input has
\[
3+3+5=11
\]
channels.

For visualization and qualitative analysis, we also define a scalar boundary-reliability map:
\[
U_{\mathrm{rel}}(u)
=
0.25U_{\mathrm{ent}}(u)
+
0.20U_{\mathrm{margin}}(u)
+
0.20U_{\mathrm{tta}}(u)
+
0.20U_{\mathrm{grad}}(u)
+
0.15U_{\mathrm{bd}}(u).
\]
The boundary-weighted reliability map is
\[
U_{\mathrm{brel}}(u)
=
U_{\mathrm{rel}}(u)
\left(0.5+0.5U_{\mathrm{bd}}(u)\right).
\]
This map is used for interpretation and visualization of uncertain boundary regions.

\subsection{Gated Residual Refinement Network}
\label{subsec:gated_refiner}

The proposed refinement module is designed to correct local errors without replacing the full base prediction. Given the 11-channel reliability tensor $R(x)$ and the base logits $\bar{z}_b$, the refiner predicts a residual logit correction $\Delta z$ and a spatial gate $G$:
\[
(\Delta z, G) = g_{\phi}\left(R(x)\right),
\]
where
\[
\Delta z \in \mathbb{R}^{C\times H\times W},
\qquad
G \in [0,1]^{1\times H\times W}.
\]
The refined logits are computed as
\[
z_r(c,u)
=
\bar{z}_b(c,u)
+
G(u)\Delta z(c,u).
\]
The refined class probabilities are then
\[
p_r(c,u)
=
\frac{\exp(z_r(c,u))}
{\sum_{k=0}^{C-1}\exp(z_r(k,u))},
\]
and the refined segmentation is
\[
\hat{y}_r(u)
=
\arg\max_{c\in\mathcal{C}}p_r(c,u).
\]

The gate $G(u)$ controls where residual corrections are applied. If $G(u)\approx 0$, the refined logits remain close to the base logits and the base prediction is preserved. If $G(u)\approx 1$, the residual correction has stronger influence. This design explicitly biases the model toward local correction rather than full segmentation replacement.

The refiner architecture is a compact encoder-decoder network. It consists of an initial convolutional block, a downsampling block, a bottleneck block, one transposed-convolution upsampling layer, a decoder block with skip connection, a $1\times1$ convolutional residual head for $\Delta z$, and a $1\times1$ sigmoid gate head for $G$. Each convolutional block contains two $3\times3$ convolutions followed by batch normalization and ReLU activation. The base channel width was set to 48.

The refiner was trained using cached TTA-mean base prediction from the training split. The refiner loss was
\[
\mathcal{L}_{\mathrm{ref}}
=
\mathcal{L}_{\mathrm{seg}}(z_r,y)
+
\lambda_g
\frac{1}{|\Omega|}
\sum_{u\in\Omega}G(u),
\]
where $\mathcal{L}_{\mathrm{seg}}$ is the compound segmentation loss defined in Section~\ref{subsec:loss}, and $\lambda_g=0.005$ is a gate regularization coefficient. The gate regularizer discourages unnecessary spatial corrections and encourages the refiner to act only where needed.

The refiner was trained with AdamW using learning rate $3\times10^{-4}$, weight decay $10^{-4}$, batch size 8, cosine learning-rate scheduling, a maximum of 80 epochs, and early stopping patience of 15 epochs. The best refiner checkpoint was selected on the validation split using
\[
S_{\mathrm{ref}}
=
\mathrm{Dice}_{\mathrm{val}}
+
0.50\,\mathrm{BoundaryDice}_{\mathrm{val}}
-
0.05\,\mathrm{ECE}_{\mathrm{val}}.
\]
This validation criterion emphasizes boundary improvement while mildly penalizing poor calibration.

\subsection{Biology-Aware Post-Processing}
\label{subsec:postprocessing}

The final segmentation was optionally post-processed using simple WBC-specific biological constraints. The post-processing assumes that WBC foreground consists of cytoplasm and nucleus, and that nucleus pixels should lie within the main cell foreground region.

Let $\hat{y}_r$ be the raw refined mask. The foreground mask is
\[
F(u)=\mathbbm{1}[\hat{y}_r(u)>0],
\]
and the nucleus mask is
\[
N(u)=\mathbbm{1}[\hat{y}_r(u)=2].
\]
First, foreground components smaller than 40 pixels were removed. The largest remaining foreground connected component was retained:
\[
F \leftarrow \mathrm{LargestComponent}
\left(
\mathrm{RemoveSmallComponents}(F,40)
\right).
\]
Foreground holes were then filled using binary hole filling. Nucleus pixels were constrained to lie inside the foreground:
\[
N \leftarrow N \cap F.
\]
Small nucleus components smaller than 15 pixels were removed:
\[
N \leftarrow \mathrm{RemoveSmallComponents}(N,15).
\]
The post-processed prediction $\tilde{y}$ is then assigned as
\[
\tilde{y}(u)
=
\begin{cases}
2, & \text{if } N(u)=1,\\
1, & \text{if } F(u)=1 \text{ and } N(u)=0,\\
0, & \text{otherwise}.
\end{cases}
\]
This step preserves multilobed nuclear structures while removing implausible small islands and enforcing foreground consistency.

\subsection{Evaluation Metrics}
\label{subsec:evaluation_metrics}

We evaluated segmentation overlap, boundary quality, probability
calibration, and selective-prediction behavior.

For class $c$, Dice and IoU are defined as
\[
\mathrm{Dice}_c
=
\frac{
2|\hat{Y}_c \cap Y_c|+\epsilon
}{
|\hat{Y}_c|+|Y_c|+\epsilon
},
\]
and
\[
\mathrm{IoU}_c
=
\frac{
|\hat{Y}_c \cap Y_c|+\epsilon
}{
|\hat{Y}_c \cup Y_c|+\epsilon
},
\]
where $\hat{Y}_c=\{u:\hat{y}(u)=c\}$ and
$Y_c=\{u:y(u)=c\}$. Mean Dice and mean IoU are computed
over the two foreground classes:
\[
\mathrm{Dice}
=
\frac{1}{C-1}
\sum_{c=1}^{C-1}
\mathrm{Dice}_c,
\qquad
\mathrm{IoU}
=
\frac{1}{C-1}
\sum_{c=1}^{C-1}
\mathrm{IoU}_c.
\]

Boundary Dice was computed by extracting binary boundaries
from the predicted and ground-truth masks for each foreground
class. For a binary mask $M$, the boundary is defined as
\[
\partial M
=
M \oplus \mathrm{Erode}(M),
\]
where $\oplus$ denotes logical exclusive-or. The Boundary
Dice for class $c$ is
\[
\mathrm{BDice}_c
=
\frac{
2|\partial \hat{Y}_c \cap \partial Y_c|+\epsilon
}{
|\partial \hat{Y}_c|+|\partial Y_c|+\epsilon
},
\]
and the reported Boundary Dice is averaged over cytoplasm
and nucleus:
\[
\mathrm{BoundaryDice}
=
\frac{1}{C-1}
\sum_{c=1}^{C-1}
\mathrm{BDice}_c.
\]

Surface-distance metrics were computed using the boundaries
of foreground classes. Let $d(a,B)$ denote the Euclidean
distance from boundary point $a$ to the closest point in
boundary set $B$. The symmetric surface-distance set is
\[
D_c
=
\{d(a,\partial Y_c): a\in\partial \hat{Y}_c\}
\cup
\{d(b,\partial \hat{Y}_c): b\in\partial Y_c\}.
\]
For each class,
\[
\mathrm{HD95}_c = \mathrm{Percentile}_{95}(D_c),
\qquad
\mathrm{ASSD}_c = \frac{1}{|D_c|}\sum_{d\in D_c} d.
\]
The reported HD95 and ASSD are averaged over the
foreground classes.

Calibration was assessed using expected calibration
error (ECE). Let
\[
\hat{p}(u)=\max_{c}p(c,u),
\qquad
\hat{y}(u)=\arg\max_c p(c,u).
\]
Pixels are partitioned into $M=15$ confidence bins
$\{B_m\}_{m=1}^{M}$. For bin $B_m$, accuracy and
confidence are
\[
\mathrm{acc}(B_m)
=
\frac{1}{|B_m|}
\sum_{u\in B_m}
\mathbbm{1}[\hat{y}(u)=y(u)],
\]
and
\[
\mathrm{conf}(B_m)
=
\frac{1}{|B_m|}
\sum_{u\in B_m}
\hat{p}(u).
\]
ECE is then
\[
\mathrm{ECE}
=
\sum_{m=1}^{M}
\frac{|B_m|}{|\Omega|}
\left|
\mathrm{acc}(B_m)-\mathrm{conf}(B_m)
\right|.
\]

For risk–coverage analysis, we used the predictive entropy \(U_{\mathrm{ent}}(u)\) defined in Section 4.4 as the ranking score, i.e., \(U_{\mathrm{rank}}(u)=U_{\mathrm{ent}}(u)\). Pixels with larger entropy were considered more uncertain. For a desired coverage level $\alpha\in(0,1]$, let
$\Omega_{\alpha}$ denote the subset containing the
$\alpha|\Omega|$ pixels with the lowest predictive
entropy. The selective risk is
\[
\mathcal{R}(\alpha)
=
\frac{1}{|\Omega_{\alpha}|}
\sum_{u\in\Omega_{\alpha}}
\mathbbm{1}[\hat{y}(u)\neq y(u)].
\]
The risk--coverage AUC was computed by numerically
integrating $\mathcal{R}(\alpha)$ over 20 uniformly
spaced coverage values from 0.05 to 1.00:
\[
\mathrm{RCAUC}
=
\frac{1}{1.00-0.05}
\int_{0.05}^{1.00}
\mathcal{R}(\alpha)\,d\alpha.
\]

\subsection{Perturbation Sensitivity of the Base Segmenter}
\label{subsec:robustness_methods}
To characterize the sensitivity of the selected base segmenter under controlled microscopy image shifts, we applied deterministic perturbations to the test images at severity levels $s\in\{0,1,2,3,4,5\}$. Severity $s=0$ corresponds to the clean image. The perturbations included brightness, contrast, gamma, blur, noise, JPEG compression, and stain-like color shift.

Brightness was adjusted by a multiplicative factor
\[
\alpha_{\mathrm{bright}} = 1 + 0.12s.
\]
Contrast was adjusted by
\[
\alpha_{\mathrm{contrast}} = 1 + 0.15s.
\]
Gamma perturbation used
\[
\gamma = \max(0.35,1-0.10s),
\]
and transformed the image intensity as
\[
x' = x^{\gamma}.
\]
Blur was implemented using Gaussian blur with radius $0.45s$. Additive Gaussian noise was applied with standard deviation $4s$ in the original 8-bit image scale. JPEG compression used quality
\[
q = \max(10,95-14s).
\]
The stain-like color shift modified RGB channels as
\[
R' = (1+0.05s)R,\qquad
G' = (1-0.025s)G,\qquad
B' = (1+0.07s)B,
\]
followed by clipping to the valid image range. Perturbation sensitivity was quantified using Dice, Boundary Dice, HD95, and ECE..

\subsection{Statistical Analysis}
\label{subsec:statistical_analysis}

To assess whether the proposed refinement produced reliable improvements over the base prediction, we performed paired image-level statistical analysis on the held-out test set. For each image $i$ and metric $m$, the paired difference was computed as
\[
\Delta_i^{(m)}
=
m_i^{\mathrm{proposed}}
-
m_i^{\mathrm{base}}.
\]
For metrics where lower values are better, such as HD95 and ECE, the sign of $\Delta_i^{(m)}$ was interpreted accordingly in the improved/worsened count.

The mean paired difference was
\[
\bar{\Delta}^{(m)}
=
\frac{1}{N}
\sum_{i=1}^{N}
\Delta_i^{(m)}.
\]
Bootstrap confidence intervals were computed using 5000 bootstrap resamples. For each resample $b$, $N$ paired differences were sampled with replacement and the mean difference was computed:
\[
\bar{\Delta}^{(m)}_b
=
\frac{1}{N}
\sum_{i\in\mathcal{I}_b}
\Delta_i^{(m)}.
\]
The 95\% confidence interval was obtained from the 2.5th and 97.5th percentiles of the bootstrap distribution:
\[
\mathrm{CI}_{95}
=
\left[
Q_{0.025}\left(\bar{\Delta}^{(m)}_b\right),
Q_{0.975}\left(\bar{\Delta}^{(m)}_b\right)
\right].
\]
In addition, a paired Wilcoxon signed-rank test was performed between the base and proposed per-image metric values. We report the mean difference, bootstrap confidence interval, Wilcoxon $p$-value, and the number of improved, worsened, and unchanged cases for each metric.

\section{Results}
\label{sec:results}

We evaluated the proposed Reliability-Aware Boundary Refinement Network (RABR-Net) for three-class white blood cell (WBC) segmentation, where the target labels correspond to background, cytoplasm, and nucleus. The evaluation was designed to assess conventional segmentation accuracy, boundary quality, probability calibration, and the spatial interpretability of the refinement process. Base-model sensitivity to controlled microscopy image perturbations was examined separately. Unless otherwise stated, all model selection and hyperparameter choices were performed using the validation set, and the test set was used only for final reporting.

\subsection{Experimental Setting and Evaluation Protocol}
\label{subsec:experimental_setting}

The primary baseline was a UNet++ architecture with an EfficientNet-B4 encoder trained at $256 \times 256$ resolution. This model was selected as the strong base segmenter because it provided a competitive balance between global overlap accuracy and boundary preservation. The proposed RABR-Net was applied as a second-stage refinement module on top of the frozen base segmenter. Given the input image, base segmentation logits, class probabilities, and uncertainty-derived reliability maps, the refiner predicts a gated residual correction rather than a full replacement segmentation. This design encourages the model to preserve confident regions while selectively correcting uncertain boundary pixels.

We report standard overlap metrics including Dice, Intersection-over-Union (IoU), and class-wise Dice for cytoplasm and nucleus, together with boundary-sensitive metrics including Boundary Dice, Hausdorff distance at the 95th percentile (HD95), and average symmetric surface distance (ASSD). Reliability is additionally examined using expected calibration error (ECE) and risk–coverage analysis. As a separate diagnostic experiment, the selected UNet++ EfficientNet-B4 base segmenter was evaluated under deterministic microscopy image perturbations.

\subsection{Main Quantitative Comparison}
\label{subsec:main_quantitative}

Table~\ref{tab:main_comparison} summarizes the main quantitative comparison across the base segmenter and the proposed refinement framework. The final proposed RABR-Net achieved the best overall Dice score, increasing the test Dice from 0.9602 for the cached TTA-mean base prediction to 0.9614 after refinement and biological post-processing. Although this global Dice improvement is modest, the improvement is more pronounced in boundary-sensitive metrics. Boundary Dice increased from 0.3448 to 0.3611, and HD95 decreased from 3.0354 to 2.8274. This indicates that the proposed refinement primarily improves spatial boundary accuracy rather than merely increasing global region overlap.

The strongest gain was observed in boundary-related performance, which is important for WBC morphology analysis because small contour errors can alter downstream measurements such as cell area, nuclear shape, cytoplasm extent, and nucleus-to-cytoplasm ratio. In contrast, ECE increased from 0.0079 to 0.0106 after refinement. This suggests that while the proposed method improves segmentation geometry, post-refinement probability calibration does not automatically improve and may require additional calibration-aware training or post-hoc temperature scaling.

\begin{table*}[t]
\centering
\caption{Main quantitative comparison on the held-out test set. Values are reported as mean \(\pm\) standard deviation over 80 test images. Higher values are better for Dice, IoU, class-wise Dice, and Boundary Dice; lower values are better for HD95, ASSD, and risk–coverage AUC. Best values are shown in bold.
}
\label{tab:main_comparison}
\scriptsize
\resizebox{\textwidth}{!}{
\begin{tabular}{lccccccccc}
\toprule
\textbf{Method} 
& \textbf{Dice} $\uparrow$
& \textbf{IoU} $\uparrow$
& \textbf{Cyto. Dice} $\uparrow$
& \textbf{Nuc. Dice} $\uparrow$
& \textbf{Boundary Dice} $\uparrow$
& \textbf{HD95} $\downarrow$
& \textbf{ASSD} $\downarrow$
& \textbf{Risk-Cov. AUC} $\downarrow$ \\
\midrule
UNet++ (Eff-B4, single-pass)
& 0.9594 $\pm$ 0.0193
& 0.9233 $\pm$ 0.0340
& 0.9471 $\pm$ 0.0314
& 0.9718 $\pm$ 0.0158
& 0.3416 $\pm$ 0.1224
& 3.0757 $\pm$ 3.4556
& 1.0449 $\pm$ 0.5825
& 0.0010 $\pm$ 0.0021 \\
Cached TTA-mean base prediction
& 0.9602 $\pm$ 0.0186
& 0.9247 $\pm$ 0.0330
& 0.9485 $\pm$ 0.0303
& 0.9720 $\pm$ 0.0156
& 0.3448 $\pm$ 0.1200
& 3.0354 $\pm$ 3.2119
& 1.0253 $\pm$ 0.5204
& 0.0009 $\pm$ 0.0015 \\
RABR-Net w/o post-processing
& 0.9613 $\pm$ 0.0187
& 0.9267 $\pm$ 0.0333
& 0.9500 $\pm$ 0.0295
& 0.9726 $\pm$ 0.0159
& 0.3609 $\pm$ 0.1308
& 2.8545 $\pm$ 2.9846
& 0.9886 $\pm$ 0.5413
& 0.0007 $\pm$ 0.0011 \\
\textbf{Proposed RABR-Net}
& \textbf{0.9614 $\pm$ 0.0186}
& \textbf{0.9268 $\pm$ 0.0332}
& \textbf{0.9501 $\pm$ 0.0294}
& \textbf{0.9727 $\pm$ 0.0158}
& \textbf{0.3611 $\pm$ 0.1308}
& \textbf{2.8274 $\pm$ 2.8682}
& \textbf{0.9821 $\pm$ 0.5114}
& \textbf{0.0007 $\pm$ 0.0011} \\
\bottomrule
\end{tabular}
}
\end{table*}


\subsection{Boundary-Focused Refinement Analysis}
\label{subsec:boundary_refinement}

\begin{figure*}[t]
    \centering
    \includegraphics[width=\linewidth]{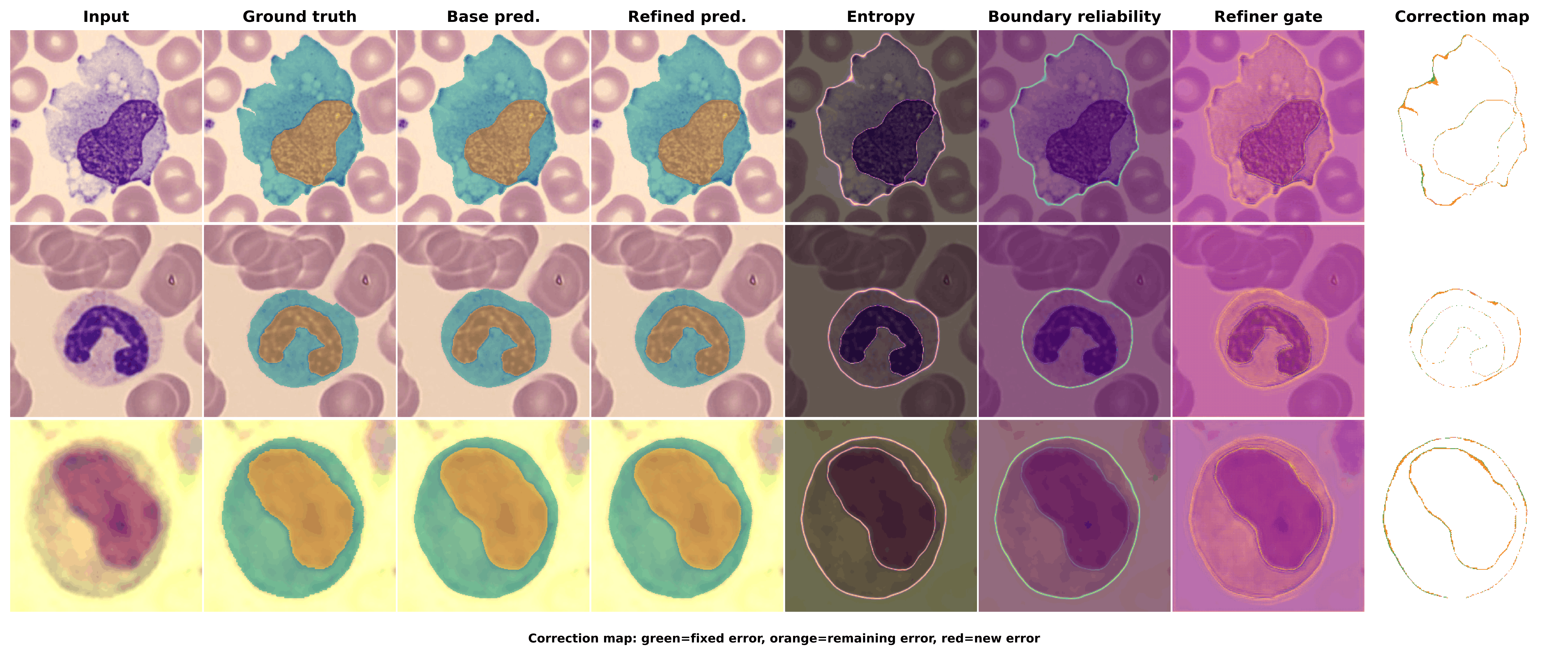}
    \caption{
    Full uncertainty-guided refinement visualization. From left to right: input image, ground truth, base prediction, refined prediction, entropy, boundary reliability, refiner gate, and correction map. The proposed method makes localized corrections near uncertain cytoplasm and nucleus boundaries. In the correction map, green denotes fixed errors, orange denotes remaining errors, and red denotes newly introduced errors.
    }
    \label{fig:uncertainty_refinement_panel}
\end{figure*}

To understand where the proposed method improves segmentation, we visualized the refinement process at the pixel level. Figure~\ref{fig:uncertainty_refinement_panel} shows a representative example containing the input image, ground truth, base prediction, refined prediction, entropy map, boundary reliability map, refiner gate, and correction map. The base and refined masks appear visually similar at the global scale, which is expected because the refiner is designed to make localized residual corrections. However, the correction map reveals that the changes are concentrated around cytoplasm and nucleus boundaries. In this example, the Dice score improves by only 0.004, while Boundary Dice improves by 0.062, demonstrating that the proposed method mainly targets boundary-level errors.

The correction map provides a direct interpretation of the refinement behavior. Green pixels indicate errors fixed by the refiner, orange pixels indicate remaining errors, and red pixels indicate newly introduced errors. The dominance of thin boundary-localized corrections supports the hypothesis that uncertainty-guided residual refinement is most beneficial in ambiguous contour regions rather than in the interior of already confident regions.

\begin{figure*}[t]
    \centering
    \includegraphics[width=\linewidth]{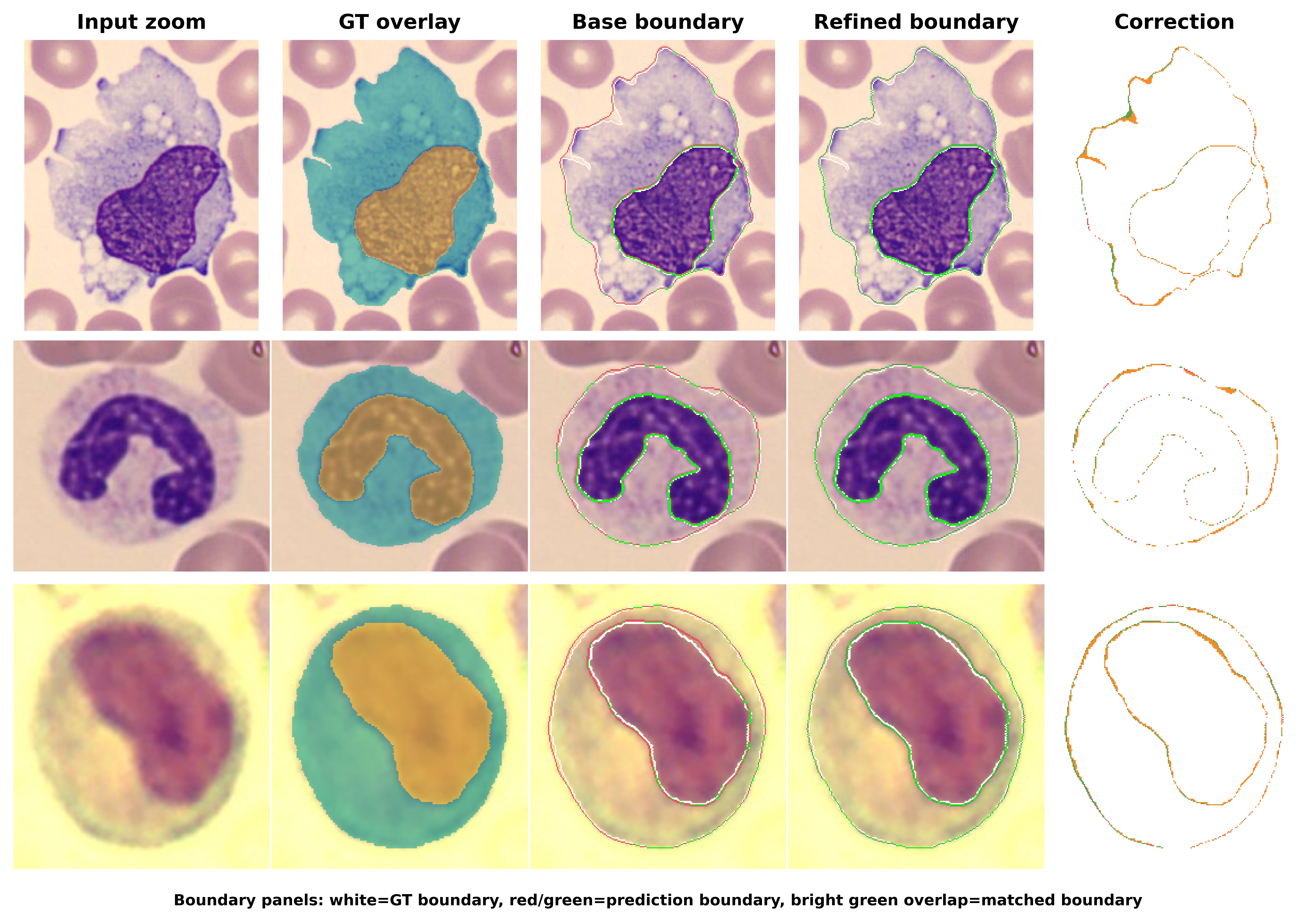}
    \caption{
    Zoomed boundary correction analysis. The proposed refiner improves localized contour errors around the cytoplasm and nucleus. Boundary overlays show the relationship between ground-truth boundaries and predicted boundaries before and after refinement.
    }
    \label{fig:boundary_zoom}
\end{figure*}

Figure~\ref{fig:boundary_zoom} further illustrates the boundary-local nature of the correction. The zoomed panels show that refinement does not substantially alter the entire cell mask. Instead, it selectively adjusts thin regions along the outer cytoplasm and inner nucleus boundaries. This behavior is desirable for WBC segmentation because biologically meaningful morphology depends on accurate delineation of both the cell boundary and nuclear contour.

\subsection{Refiner Gate and Correction Mechanism}
\label{subsec:gate_mechanism}

\begin{figure*}[t]
    \centering
    \begin{subfigure}{\linewidth}
        \centering
        \includegraphics[width=0.45\linewidth]{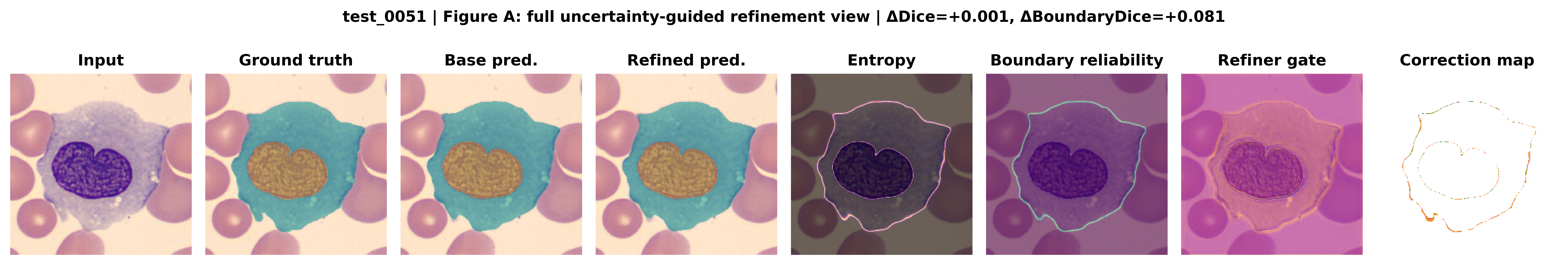}
        \caption{Full context of uncertainty-guided refinement.}
    \end{subfigure}
    \vspace{0.4em}
    \begin{subfigure}{\linewidth}
        \centering
        \includegraphics[width=0.45\linewidth]{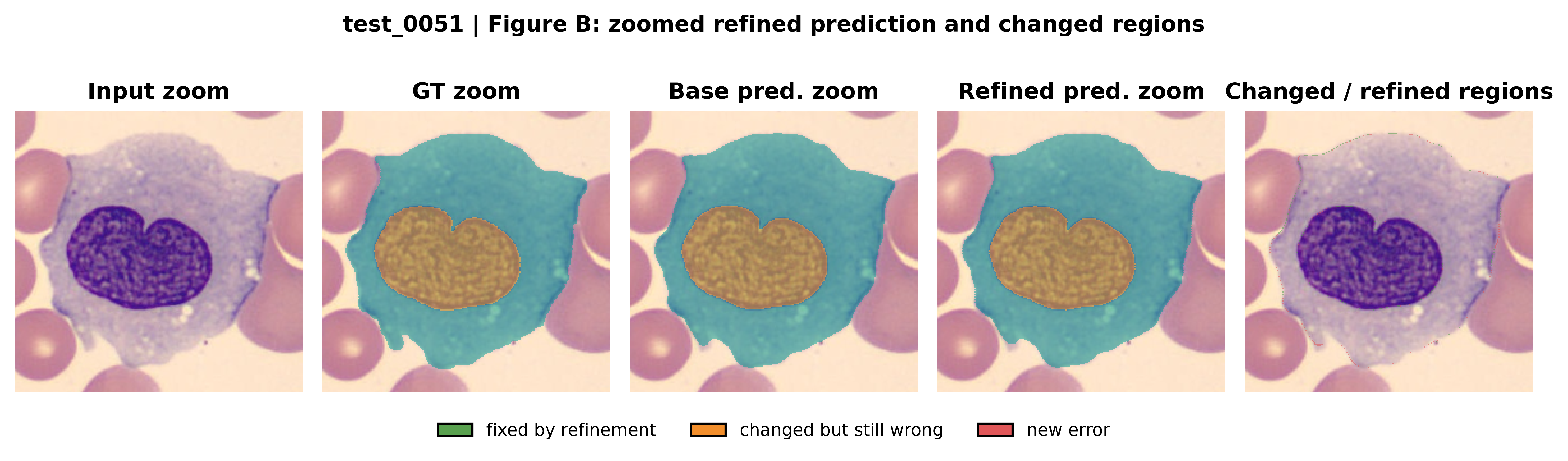}
        \caption{Zoomed refined prediction and changed regions.}
    \end{subfigure}
    \vspace{0.4em}
    \begin{subfigure}{\linewidth}
        \centering
        \includegraphics[width=0.45\linewidth]{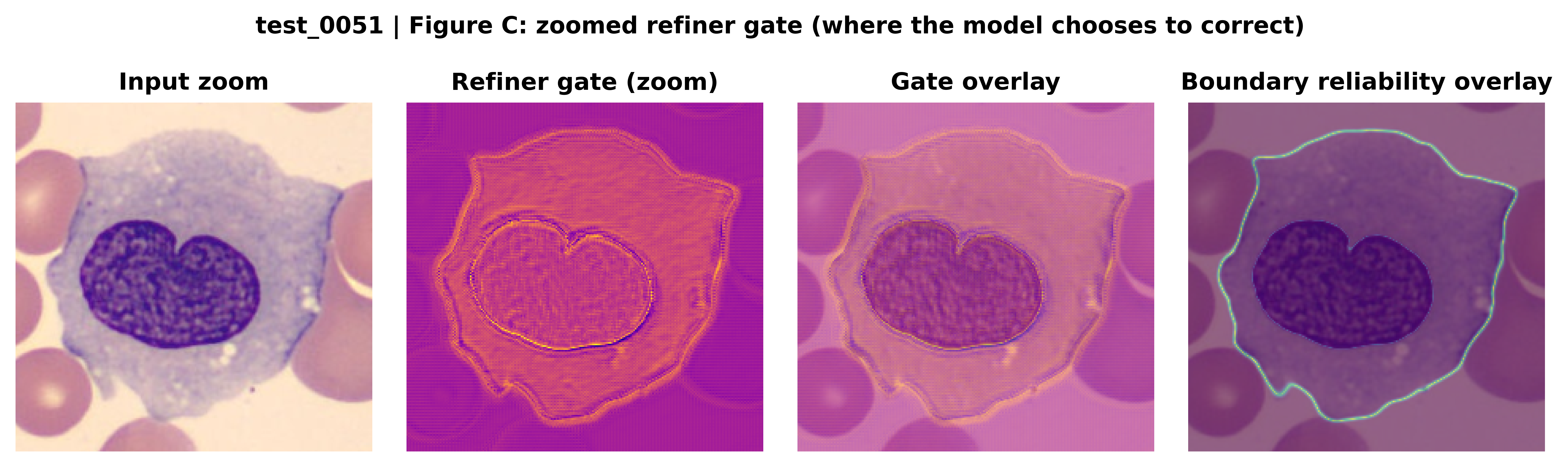}
        \caption{Zoomed refiner gate.}
    \end{subfigure}
    \hfill
    \begin{subfigure}{\linewidth}
        \centering
        \includegraphics[width=0.45\linewidth]{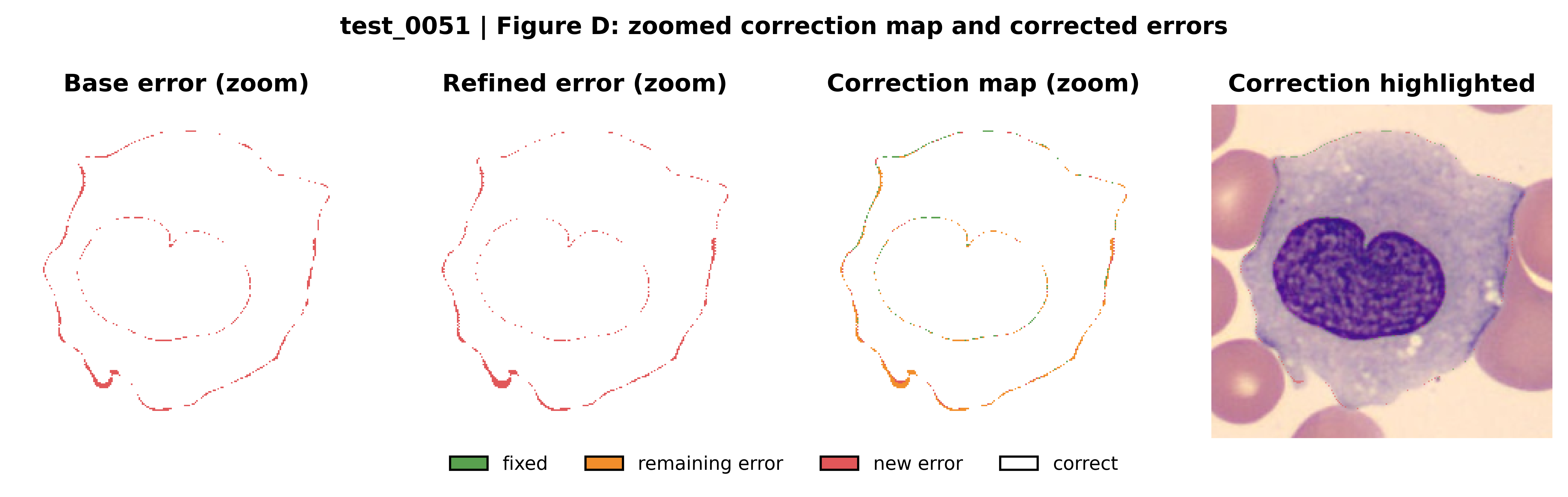}
        \caption{Zoomed correction map.}
    \end{subfigure}
    \caption{
    Case study of the proposed gated residual refinement mechanism. The refiner gate is concentrated around uncertain cytoplasm and nucleus boundaries, and the correction map shows which errors are fixed, remain unresolved, or are newly introduced.
    }
    \label{fig:case_study_gate_correction}
\end{figure*}

The proposed refiner uses a gating mechanism to control where residual corrections are applied. Figure~\ref{fig:case_study_gate_correction} shows a detailed case study divided into four complementary views. Figure~\ref{fig:case_study_gate_correction}A presents the complete refinement pipeline. Figure~\ref{fig:case_study_gate_correction}B shows zoomed refined predictions and highlights the pixels changed by refinement. Figure~\ref{fig:case_study_gate_correction}C visualizes the refiner gate, showing that the gate is activated primarily around uncertain cytoplasm and nucleus boundaries. Figure~\ref{fig:case_study_gate_correction}D shows the zoomed correction map, separating fixed errors, remaining errors, and newly introduced errors.

These visualizations support the intended mechanism of RABR-Net. The refiner does not behave as an unconstrained second segmentation network. Instead, it acts as a reliability-guided correction module that uses uncertainty and boundary evidence to decide where local residual updates should be applied. This improves interpretability because the correction process can be spatially inspected and compared with uncertainty maps and segmentation errors.

%

\subsection{Uncertainty and Reliability Maps Localize Ambiguous Boundaries}
\label{subsec:uncertainty_reliability}
We next examined the uncertainty and boundary-related source maps used by the refinement pipeline, as illustrated in Fig. 5. Entropy highlights low-confidence regions in the base prediction, TTA variance captures prediction instability under test-time transformations, and margin uncertainty identifies pixels where competing class probabilities are difficult to separate. The foreground probability-gradient and soft-boundary terms primarily emphasize transitions between the combined cell foreground and background. In contrast, the multiclass probabilities and uncertainty measures can also respond to ambiguity at the internal nucleus–cytoplasm interface.

Qualitatively, the combined information is concentrated around morphologically relevant transition regions rather than homogeneous cell interiors. The learned refiner gate shows a similar spatial pattern, indicating that the network tends to apply corrections near ambiguous contours.

\begin{figure*}[t]
    \centering
    \includegraphics[width=\linewidth]{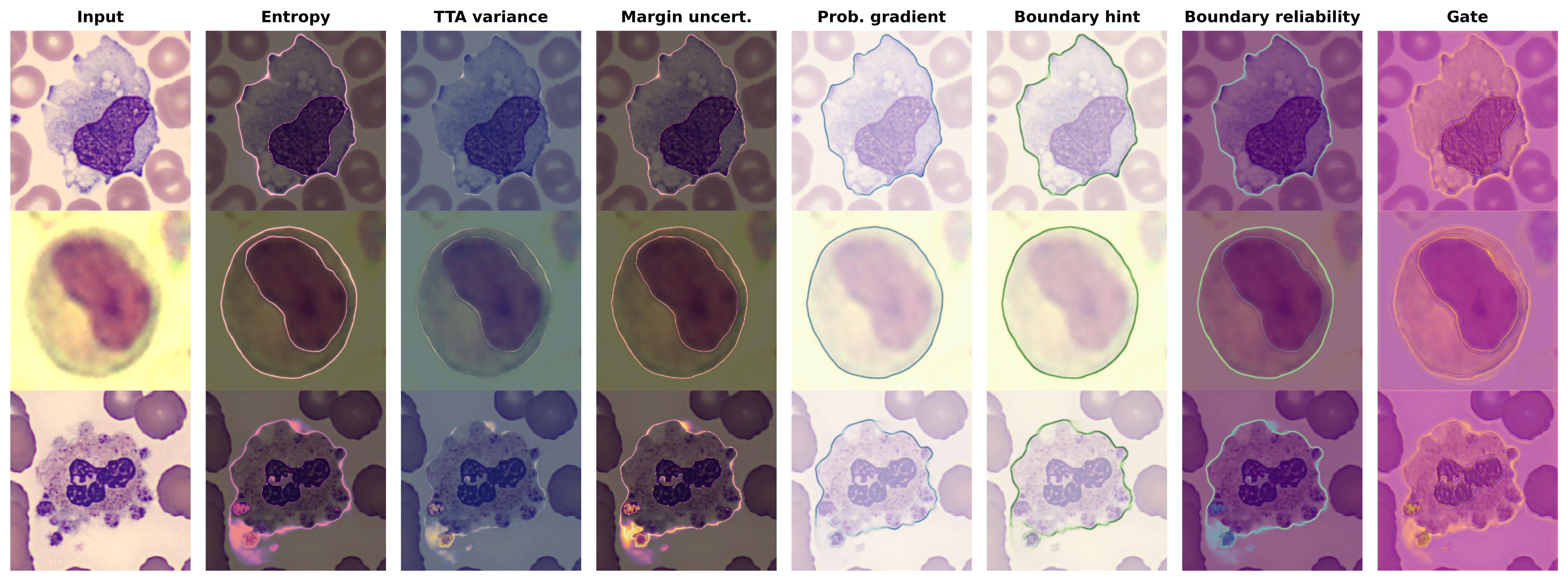}
    \caption{Uncertainty and reliability-source maps used in the refinement pipeline. Entropy, TTA variance, and margin uncertainty characterize predictive ambiguity, whereas the foreground probability-gradient and soft-boundary terms emphasize the outer cell contour. Together with the multiclass probability channels, these signals provide complementary information for localized refinement. The learned refiner gate is concentrated predominantly around morphologically relevant boundary regions.
    }
    \label{fig:uncertainty_gallery}
\end{figure*}

\subsection{Perturbation Sensitivity of the Base Segmenter}
\label{subsec:robustness}
To characterize the sensitivity of the selected base segmenter to realistic image-quality shifts, we evaluated the UNet++ EfficientNet-B4 model under brightness variation, contrast change, gamma shift, blur, additive noise, JPEG compression, and stain-like color shift. These perturbations simulate common sources of variability in microscopy images, including differences in staining, illumination, focus, compression, and acquisition conditions.

Table~\ref{tab:robustness} summarizes the robustness results. The model remained relatively stable under blur, noise, and JPEG compression, with only modest degradation across severity levels. In contrast, contrast shift, brightness shift, and stain-like color variation produced larger performance drops at high severity. For example, contrast severity 5 reduced Dice to 0.8815 and Boundary Dice to 0.2588, while increasing HD95 to 8.2696. These findings indicate that color and intensity shifts remain important failure modes for reliable microscopy segmentation.

These experiments characterize the sensitivity of the selected base model to acquisition-related perturbations. Because the perturbation study was not performed on the complete refinement pipeline, the results should be interpreted as a diagnostic analysis of backbone sensitivity rather than as evidence of robustness of RABR-Net as a whole.

\begin{table*}[!ht]
\centering
\caption{
Robustness summary across perturbation types and severity levels for the selected UNet++ EfficientNet-B4 model. Severity 0 denotes the clean condition. Values are means over 80 test images.
}
\label{tab:robustness}
\scriptsize
\resizebox{0.70\textwidth}{!}{
\begin{tabular}{llcccc}
\toprule
\textbf{Perturbation} 
& \textbf{Severity}
& \textbf{Dice} $\uparrow$
& \textbf{Boundary Dice} $\uparrow$
& \textbf{HD95} $\downarrow$
& \textbf{ECE} $\downarrow$ \\
\midrule
Brightness & 0 & 0.9594 & 0.3416 & 3.0757 & 0.0093 \\
Brightness & 1 & 0.9579 & 0.3424 & 3.1904 & 0.0101 \\
Brightness & 2 & 0.9539 & 0.3320 & 3.3361 & 0.0117 \\
Brightness & 3 & 0.9470 & 0.3164 & 4.0098 & 0.0142 \\
Brightness & 4 & 0.9370 & 0.3049 & 4.8061 & 0.0173 \\
Brightness & 5 & 0.9224 & 0.2781 & 5.7344 & 0.0224 \\
\midrule
Contrast & 0 & 0.9594 & 0.3416 & 3.0757 & 0.0093 \\
Contrast & 1 & 0.9586 & 0.3381 & 3.1371 & 0.0094 \\
Contrast & 2 & 0.9529 & 0.3232 & 3.7057 & 0.0105 \\
Contrast & 3 & 0.9419 & 0.3066 & 4.6014 & 0.0120 \\
Contrast & 4 & 0.9126 & 0.2809 & 6.6516 & 0.0177 \\
Contrast & 5 & 0.8815 & 0.2588 & 8.2696 & 0.0233 \\
\midrule
Gamma & 0 & 0.9594 & 0.3416 & 3.0757 & 0.0093 \\
Gamma & 1 & 0.9590 & 0.3396 & 3.1283 & 0.0095 \\
Gamma & 2 & 0.9574 & 0.3299 & 3.0695 & 0.0099 \\
Gamma & 3 & 0.9540 & 0.3153 & 3.2307 & 0.0107 \\
Gamma & 4 & 0.9491 & 0.2970 & 3.6131 & 0.0118 \\
Gamma & 5 & 0.9426 & 0.2706 & 3.9738 & 0.0131 \\
\midrule
Blur & 0 & 0.9594 & 0.3416 & 3.0757 & 0.0093 \\
Blur & 1 & 0.9594 & 0.3415 & 3.0801 & 0.0093 \\
Blur & 2 & 0.9594 & 0.3417 & 3.0634 & 0.0093 \\
Blur & 3 & 0.9593 & 0.3415 & 3.0456 & 0.0092 \\
Blur & 4 & 0.9592 & 0.3405 & 3.0333 & 0.0092 \\
Blur & 5 & 0.9586 & 0.3340 & 3.0311 & 0.0094 \\
\midrule
Noise & 0 & 0.9594 & 0.3416 & 3.0757 & 0.0093 \\
Noise & 1 & 0.9588 & 0.3398 & 3.1904 & 0.0096 \\
Noise & 2 & 0.9584 & 0.3379 & 3.1759 & 0.0095 \\
Noise & 3 & 0.9574 & 0.3358 & 3.2451 & 0.0096 \\
Noise & 4 & 0.9568 & 0.3375 & 3.2467 & 0.0100 \\
Noise & 5 & 0.9566 & 0.3379 & 3.3458 & 0.0100 \\
\midrule
JPEG & 0 & 0.9594 & 0.3416 & 3.0757 & 0.0093 \\
JPEG & 1 & 0.9593 & 0.3396 & 3.0818 & 0.0093 \\
JPEG & 2 & 0.9590 & 0.3393 & 3.1254 & 0.0094 \\
JPEG & 3 & 0.9591 & 0.3402 & 3.0506 & 0.0093 \\
JPEG & 4 & 0.9589 & 0.3386 & 3.0947 & 0.0093 \\
JPEG & 5 & 0.9585 & 0.3355 & 3.1060 & 0.0092 \\
\midrule
Stain shift & 0 & 0.9594 & 0.3416 & 3.0757 & 0.0093 \\
Stain shift & 1 & 0.9580 & 0.3349 & 3.2220 & 0.0099 \\
Stain shift & 2 & 0.9559 & 0.3225 & 3.4044 & 0.0109 \\
Stain shift & 3 & 0.9529 & 0.3059 & 3.7626 & 0.0121 \\
Stain shift & 4 & 0.9484 & 0.2829 & 4.2930 & 0.0134 \\
Stain shift & 5 & 0.9422 & 0.2559 & 5.8902 & 0.0153 \\
\bottomrule
\end{tabular}
}
\end{table*}

\subsection{Ablation Study}
\label{subsec:ablation}
We performed a pipeline-stage ablation to quantify the contribution of the main refinement components. Starting from the selected UNet++ EfficientNet-B4 base segmenter, we first report the cached TTA-mean base prediction used by the refinement pipeline. We then evaluate the reliability-guided gated residual refiner and finally the addition of biological post-processing.

Table~\ref{tab:ablation} shows that the largest improvement occurs after introducing the reliability-guided gated residual refiner. Boundary Dice increases from 0.3448 for the cached TTA-mean base prediction to 0.3609 after refinement, while HD95 decreases from 3.0354 to 2.8545. Biological post-processing provides a smaller additional improvement in Dice, Boundary Dice, and HD95, producing the final proposed result. These findings indicate that, among the pipeline stages evaluated here, the largest performance change occurs after introduction of the reliability-guided gated residual refinement stage, while biological post-processing provides a smaller additional structural-consistency benefit.

A more granular ablation of individual reliability-map components, such as entropy, TTA variance, margin uncertainty, probability gradients, and boundary cues, remains an important direction for future analysis.

\begin{table*}[h]
\centering
\caption{
Pipeline-stage ablation on the held-out test set. Each row adds one stage of the final pipeline. The largest observed gain occurs after addition of the reliability-guided gated residual refiner, whereas biological post-processing provides a smaller additional improvement.
}
\label{tab:ablation}
\scriptsize
\resizebox{\textwidth}{!}{
\begin{tabular}{lccccccc}
\toprule
\textbf{Configuration}
& \textbf{Dice} $\uparrow$
& \textbf{IoU} $\uparrow$
& \textbf{Boundary Dice} $\uparrow$
& \textbf{$\Delta$ Boundary Dice}
& \textbf{HD95} $\downarrow$
& \textbf{Risk-Cov. AUC} $\downarrow$ \\
\midrule
Base UNet++ (Eff-B4, single-pass)
& 0.9594
& 0.9233
& 0.3416
& --
& 3.0757
& 0.0010 \\
Cached TTA-mean base prediction
& 0.9602
& 0.9247
& 0.3448
& +0.0032
& 3.0354
& 0.0009 \\
+ Reliability-guided gated residual refiner
& 0.9613
& 0.9267
& 0.3609
& +0.0161
& 2.8545
& 0.0007 \\
\textbf{+ Biological post-processing}
& \textbf{0.9614}
& \textbf{0.9268}
& \textbf{0.3611}
& +0.0002
& \textbf{2.8274}
& \textbf{0.0007} \\
\bottomrule
\end{tabular}
}
\end{table*}

\subsection{Statistical Reliability of the Improvement}
\label{subsec:statistics}

Because the base segmenter already achieves high global Dice, small absolute improvements in Dice may still correspond to meaningful boundary corrections. Therefore, we evaluated statistical reliability using paired per-image comparisons between the base and refined predictions. For each metric, we computed bootstrap confidence intervals and paired significance tests. We also counted the number of test images for which the proposed method improved, degraded, or remained approximately unchanged.

This analysis is particularly important for Boundary Dice and HD95, where the proposed method is expected to provide its most meaningful gains. Reporting per-image improvement statistics helps distinguish systematic boundary correction from isolated improvements in a small number of examples. The paired image-level statistical comparisons are summarized in Table \ref{tab:statistical_reliability}.

\begin{table*}[t]
\centering
\caption{Paired image-level statistical comparison of the final RABR-Net segmentation with the cached TTA-mean base prediction on the held-out test set. The reported difference is \(\Delta=\mathrm{Proposed}-\mathrm{Base}\). Positive \(\Delta\) indicates improvement for Dice and Boundary Dice, whereas negative \(\Delta\) indicates improvement for HD95. Improved/worsened/equal counts are direction-aware.
}
\label{tab:statistical_reliability}
\scriptsize
\resizebox{\textwidth}{!}{
\begin{tabular}{lcccccc}
\toprule
\textbf{Metric}
& \textbf{Base Mean}
& \textbf{Proposed Mean}
& \textbf{Mean $\Delta$}
& \textbf{95\% Bootstrap CI}
& \textbf{Wilcoxon $p$}
& \textbf{Improved/Worsened/Equal} \\
\midrule
Dice $\uparrow$
& 0.9602
& 0.9614
& +0.0011
& [0.0005, 0.0018]
& $3.66 \times 10^{-4}$
& 55 / 25 / 0 \\
Boundary Dice $\uparrow$
& 0.3448
& 0.3611
& +0.0163
& [0.0083, 0.0247]
& $1.17 \times 10^{-3}$
& 52 / 28 / 0 \\
HD95 $\downarrow$
& 3.0354
& 2.8274
& -0.2080
& [-0.4127, -0.0720]
& $1.62 \times 10^{-3}$
& 37 / 15 / 28 \\
\bottomrule
\end{tabular}
}
\end{table*}
\subsection{Failure Cases and Limitations Observed in the Results}
\label{subsec:failure_cases}

Finally, we inspected failure and low-improvement cases to better understand the limitations of the proposed method. Failure cases generally occur when uncertainty does not align well with the true segmentation error, when the base model is confidently wrong, or when image degradation obscures the cytoplasm or nucleus boundary. In such cases, the refiner may leave some errors unresolved or introduce small new boundary errors.

\begin{figure*}[t]
    \centering
    \includegraphics[width=\linewidth]{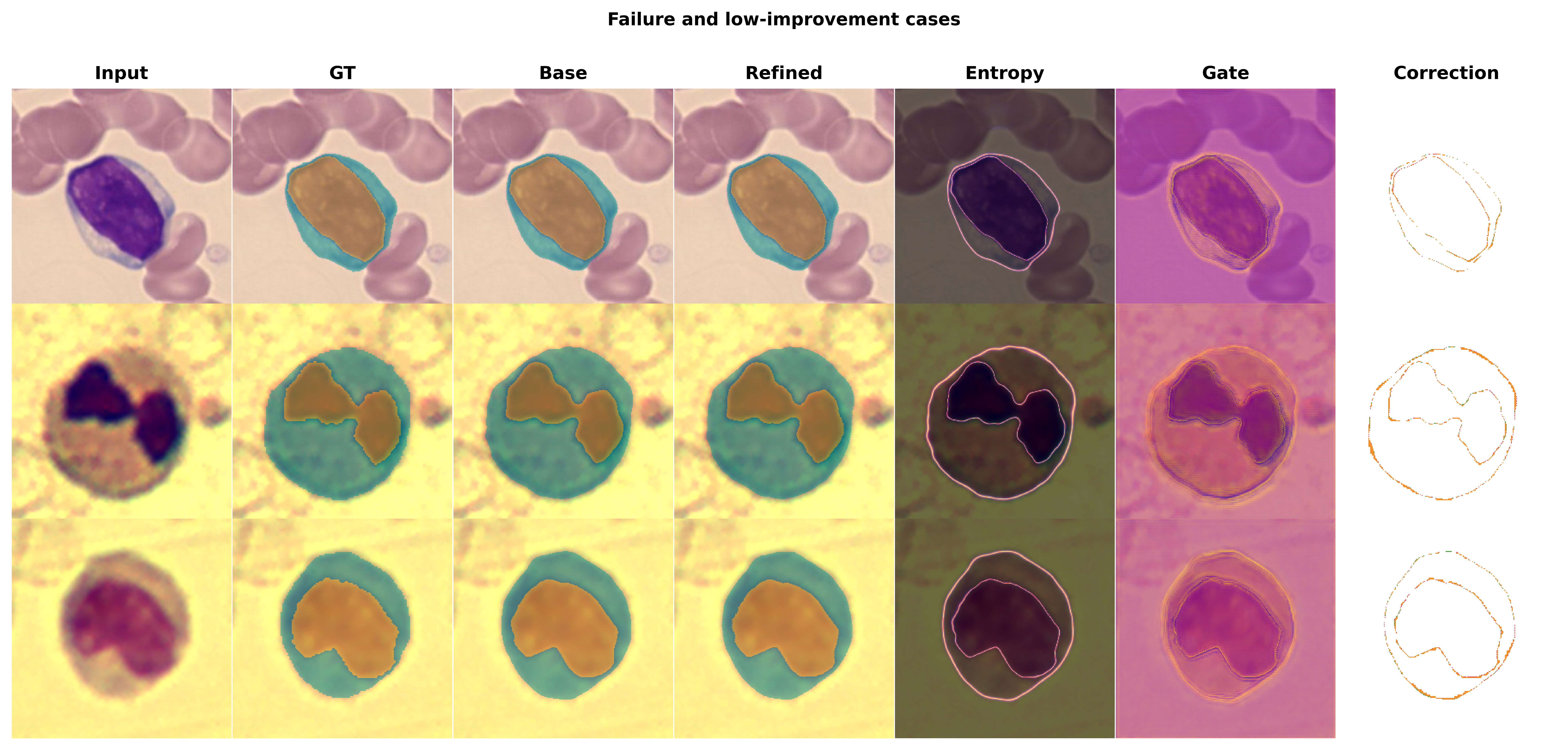}
    \caption{
    Failure and low-improvement examples. These cases show that refinement is most effective when uncertainty overlaps with true boundary error. When the base segmenter is confidently wrong or image quality is poor, some errors may remain unresolved or new small errors may be introduced.
    }
    \label{fig:failure_cases}
\end{figure*}

This observation highlights an important distinction between spatial correction and probability calibration. Although residual refinement improves boundary geometry, the corresponding probabilistic output is not better calibrated. Calibration-aware objectives or post-hoc recalibration therefore remain potential extensions of the present framework. Representative failure and low-improvement cases are shown in Fig. \ref{fig:failure_cases}

\subsection{Summary of Findings}
\label{subsec:results_summary}

Overall, the results show that RABR-Net improves WBC segmentation primarily by correcting uncertain boundary regions rather than by globally changing the segmentation mask. The proposed method achieves the best Dice, Boundary Dice, and HD95 among the compared settings, with the most meaningful gains observed in boundary-sensitive metrics. Qualitative analyses further confirm that the refiner gate is concentrated near cytoplasm and nucleus boundaries and that many corrected pixels correspond to biologically meaningful contour errors. However, calibration metrics indicate that improved spatial accuracy does not necessarily imply improved confidence calibration, motivating future work on calibration-aware reliability refinement.

\section{Discussion}
\label{sec:discussion}

This study introduced RABR-Net, a reliability-aware boundary refinement framework for trustworthy biomedical image segmentation. The central motivation was that high global segmentation accuracy alone is insufficient when the downstream task depends on precise morphology. In blood-smear image segmentation, cytoplasm and nucleus boundaries define biologically meaningful measurements such as cell area, nuclear contour, cytoplasm extent, and nucleus-to-cytoplasm ratio. Therefore, a model that improves boundary geometry, even with a modest change in global Dice, can provide a more useful segmentation for morphology-sensitive analysis.

\subsection{Boundary-Sensitive Improvement}
\label{subsec:discussion_boundary}

The strongest effect of the proposed method was observed in boundary-sensitive metrics. Compared with the cached TTA-mean base prediction, RABR-Net improved Boundary Dice from 0.3448 to 0.3611 and reduced HD95 from 3.0354 to 2.8274 on the held-out test set. These improvements are important because Boundary Dice directly measures agreement between predicted and true object contours, while HD95 captures large boundary deviations and outlier surface errors. In microscopy segmentation, such errors may occur at the cytoplasm-background interface or around irregular nuclear contours. These regions occupy only a small fraction of the image, but they can strongly influence downstream morphological measurements.

The improvement in HD95 is particularly meaningful because it indicates that the proposed refinement reduces extreme boundary deviations. A lower HD95 suggests that the refined segmentation is not only slightly better on average, but also less likely to contain large contour errors. This is consistent with the design of RABR-Net: the method does not attempt to replace the full segmentation mask, but instead applies local residual corrections in uncertain boundary regions.

\subsection{Why the Global Dice Gain Is Small but Meaningful}
\label{subsec:discussion_dice}

The absolute gain in global Dice was modest, increasing from 0.9602 for the cached TTA-mean base prediction to 0.9614 after the proposed post-processed refinement. This small numerical gain should be interpreted in the context of a very strong baseline. When the base model already segments most foreground pixels correctly, global Dice becomes dominated by large correctly segmented interior regions. As a result, improvements localized to thin boundary regions may produce only small changes in Dice.

However, the combination of improved Dice, improved Boundary Dice, and reduced HD95 shows that the refinement is not random or cosmetic. The paired statistical analysis further supports this interpretation: RABR-Net significantly improved Dice, Boundary Dice, and HD95 compared with the base prediction. Therefore, the contribution of the proposed method is best understood as morphology-preserving boundary correction rather than large-scale mask replacement.

\subsection{Calibration Does Not Automatically Improve After Refinement}
\label{subsec:discussion_calibration}

Probability calibration was evaluated on model outputs before biological post-processing. ECE increased from 0.0079 for the cached TTA-mean base prediction to 0.0106 after residual refinement. Thus, although the refiner improved segmentation and boundary geometry, it did not improve probability calibration. This provides a more specific interpretation of the method: spatial segmentation quality and probabilistic confidence quality are related but distinct properties.

This behavior is plausible because the refinement objective primarily optimizes segmentation and boundary-related accuracy rather than calibration. The gated residual correction modifies logits in uncertain regions and may improve the resulting segmentation mask while simultaneously altering the confidence distribution.

The subsequent biology-aware post-processing operates on the discrete segmentation mask rather than constituting a probabilistic calibration operation. We therefore do not interpret post-processing as an additional calibration stage. Calibration-aware training objectives or validation-based post-hoc calibration methods could be investigated separately in future work.


\subsection{Interpretability Through Uncertainty and Correction Maps}
\label{subsec:discussion_interpretability}

A key advantage of the proposed framework is that the
refinement process can be visually inspected. The uncertainty
maps, boundary reliability maps, refiner gates, and correction
maps provide complementary views of how the model behaves.
The uncertainty sources provide complementary information:
entropy and margin uncertainty capture class ambiguity,
TTA variance reflects transformation-induced prediction
instability, and foreground probability gradients emphasize
spatial changes at the outer WBC contour. Together with the
multiclass probability channels, these signals allow the
refiner to identify regions in which localized corrections
may be useful, including ambiguity around the
nucleus--cytoplasm interface.

The learned gate provides an interpretable mechanism for
refinement. Qualitative results show that the gate is
concentrated around cytoplasm and nucleus boundaries rather
than uniformly across the whole image. This supports the
intended behavior of the method: RABR-Net selectively
corrects uncertain boundary pixels while preserving
confident interior regions. Correction maps further reveal
which pixels are fixed, which errors remain unresolved,
and whether new errors are introduced. This makes the
refinement process more transparent than a conventional
second-stage segmentation model.

\subsection{Clinical and Biological Relevance}
\label{subsec:discussion_clinical}

The proposed method is relevant to morphology-sensitive biomedical image analysis because it improves the regions that are most important for structural interpretation. In WBC microscopy, accurate cytoplasm and nucleus boundaries are necessary for estimating nuclear shape, cytoplasmic area, and nucleus-to-cytoplasm ratio. These features are commonly associated with downstream cell characterization and abnormality screening workflows.

The framework should be understood as a segmentation-support method rather than a diagnostic system. It does not make clinical decisions directly. Instead, it aims to produce more reliable segmentation masks that may support downstream quantitative morphology analysis. By exposing uncertainty maps and correction maps, the framework may also help users identify regions where the segmentation should be inspected more carefully.

\subsection{Sensitivity of the Base Segmenter to Image Perturbations}
\label{subsec:discussion_robustness}

The perturbation analysis characterizes the selected UNet++ EfficientNet-B4 base segmenter rather than the complete RABR-Net pipeline. The base model remained relatively stable under blur, noise, and JPEG compression, whereas stronger contrast, brightness, and stain-like shifts produced larger degradation. These observations identify color and intensity variation as important sensitivities of the underlying segmentation backbone.

Because refinement was not evaluated under these perturbation conditions, these results should not be interpreted as demonstrating robustness of the complete RABR-Net framework. Instead, they provide complementary information about the behavior of the base model from which the refinement pipeline starts.

\subsection{Limitations}
\label{subsec:discussion_limitations}

This study has several limitations. First, the evaluation
was performed on a single benchmark dataset with a fixed
train/validation/test split. Although the held-out test set
was preserved for final reporting, external validation on
independent microscopy datasets is necessary to assess
generalization across staining protocols, acquisition
devices, and laboratory conditions.

Second, although residual refinement improves segmentation
geometry, these improvements do not necessarily translate
into better probability calibration. In our evaluation,
ECE increases for the probabilistic refined output before
post-processing relative to the cached TTA-mean base prediction.
Segmentation accuracy and probabilistic calibration should
therefore be treated as distinct objectives. Future work
may investigate calibration-aware losses or post-hoc
recalibration to address this limitation.

Third, the refiner depends on the quality of the base
model. If the base model is confidently incorrect,
uncertainty cues may not reliably identify the affected
regions, limiting the refiner's ability to correct such
errors. Qualitative failure cases illustrate that
uncertainty does not always coincide with segmentation
errors. Furthermore, the perturbation analysis was
restricted to the selected base segmenter. The robustness
of the complete RABR-Net pipeline under acquisition and
staining shifts therefore remains to be established.

Fourth, biological post-processing relies on simple
structural assumptions, including preservation of the
main foreground component and containment of nucleus
pixels within the cell foreground. Although appropriate
for the present segmentation setting, these assumptions
may not generalize to overlapping cells or more complex
cellular morphologies, for which more flexible
topology-aware constraints may be required.

Finally, although qualitative uncertainty and correction
maps provide useful insights into the refinement process,
they do not replace expert validation. Future studies
should investigate whether the corrected segmentation
masks improve downstream morphological measurements and
whether domain experts find the uncertainty and
correction maps informative for practical analysis.

\subsection{Future Work}
\label{subsec:discussion_future_work}

Future work will focus on four main directions. First, external validation should be performed on independent blood-smear microscopy datasets acquired under different staining, scanning, and laboratory conditions. This would provide a stronger assessment of cross-domain generalization.

Second, calibration-aware refinement should be incorporated into the training objective. Possible approaches include temperature scaling, differentiable ECE surrogates, Brier-score regularization, uncertainty-consistency losses, and conformal risk-control strategies. These methods may help improve both mask quality and confidence reliability.

Third, the biological constraints can be extended using topology-aware or shape-aware learning. Instead of applying post-processing only after prediction, future models could incorporate differentiable topology losses, contour priors, or nucleus-cytoplasm containment constraints during training.

Fourth, the framework should be evaluated on downstream morphology tasks. For example, future work could test whether RABR-Net improves the accuracy of nuclear area, cytoplasmic area, contour irregularity, and nucleus-to-cytoplasm ratio measurements. Such analysis would directly connect boundary refinement with practical biomedical image analysis outcomes.
\section{Conclusion}
\label{sec:conclusion}

This work presented RABR-Net, a reliability-aware boundary refinement framework for biomedical image segmentation. The method combines a strong base segmenter with predictive uncertainty information, foreground-derived boundary cues, gated residual refinement, and biology-aware post-processing. Rather than replacing the complete segmentation mask, the refiner predicts localized corrections while preserving regions that do not require modification.

On the held-out WbcMSBench test set, the complete pipeline improved Dice, Boundary Dice, and HD95 relative to the cached TTA-mean base prediction. The largest observed improvement across the evaluated pipeline stages occurred after introduction of the gated residual refinement stage, while biological post-processing provided a smaller additional improvement. Qualitative visualizations further showed that the learned gate and resulting corrections are concentrated predominantly around cytoplasm and nucleus contours.

Probability calibration did not improve after residual refinement, demonstrating that improved segmentation geometry does not necessarily imply improved confidence calibration. A separate perturbation analysis of the base segmenter also identified sensitivity to strong color and intensity shifts. Taken together, the results support evaluating morphology-sensitive biomedical segmentation using complementary overlap, boundary, calibration, selective-risk, qualitative, and paired statistical analyses rather than global Dice alone.
\bibliographystyle{splncs04}
\bibliography{ref}
\end{document}